\documentclass[]{article}

\usepackage{times}
\usepackage{epsfig}
\usepackage{graphicx}
\usepackage{amsmath}
\usepackage{amssymb}
\usepackage{multirow}
\usepackage{booktabs}
\usepackage{color}
\usepackage{footnote}
\usepackage{engord}

\usepackage{algorithm}
\usepackage{algorithmic}
\usepackage{amsmath}
\usepackage{amsfonts}
\usepackage{mathtools}
\usepackage{booktabs}

\include{defs}

\usepackage{hyperref}

\usepackage{marvosym}
\usepackage{lipsum}

\usepackage{appendix}

\newtheorem{definition}{Definition}

\newcommand{\revise}[1]{{\color{red}#1}}

\newcommand{\executeiffilenewer}[3]{%
	\ifnum\pdfstrcmp{\pdffilemoddate{#1}}%
	{\pdffilemoddate{#2}}>0%
	{\immediate\write18{#3}}\fi%
}

\graphicspath{{./}{section/figure/}{images/}}

\title{Transferability Estimation for Semantic Segmentation Task}
\author{Yang Tan, Yang Li, Shao-Lun Huang \\ 
Tsinghua-Berkeley Shenzhen Institute, Tsinghua University \\
\footnotesize{tany19@mails.tsinghua.edu.cn, \{yangli, shaolun.huang\}@sz.tsinghua.edu.cn}
}

\begin{document}

\maketitle

\revise{This paper is not the latest version. Please refer to \href{https://ieeexplore.ieee.org/abstract/document/10222912}{\textit{Efficient Prediction of Model Transferability in Semantic Segmentation Tasks (ICIP'23)}} for more details.}

\begin{abstract}
Transferability estimation is a fundamental problem in transfer learning to predict how good the performance is when transferring a source model (or source task) to a target task. With the guidance of transferability score, we can efficiently select the highly transferable source models without performing the real transfer in practice. Recent analytical transferability metrics are mainly designed for image classification problem, and currently there is no specific investigation for the transferability estimation of semantic segmentation task, which is an essential problem in autonomous driving, medical image analysis, etc. Consequently, we further extend the recent analytical transferability metric OTCE (Optimal Transport based Conditional Entropy) score to the semantic segmentation task. The challenge in applying the OTCE score is the high dimensional segmentation output, which is difficult to find the optimal coupling between so many pixels under an acceptable computation cost. Thus we propose to randomly sample N pixels for computing OTCE score and take the expectation over K repetitions as the final transferability score. Experimental evaluation on Cityscapes, BDD100K and GTA5 datasets demonstrates that the OTCE score highly correlates with the transfer performance. 
 
\end{abstract}

\section{Introduction}
Transfer learning is quite useful in improving the performance of few-labeled tasks with the help of related source tasks (or source models)~\cite{pratt1993discriminability, sun2019meta}. Consequently, understanding the relationship between source and target tasks is crucial to the success of transfer learning. However, characterizing the task relatedness is a nontrivial problem since empirically evaluating the transfer performance of each task pair is computational expensive and time consuming.  Recent analytical transferability metrics~\cite{NCE,bao2019information,LEEP,OTCE} are able to assess the transferability of a source model in a more efficient manner. In practice, we can not only apply transferability metrics to select the best source model, but also help ranking the highly transferable tasks for joint training~\cite{zamir2018taskonomy} and multi-source feature fusion~\cite{OTCE}. Semantic segmentation is a fundamental visual task for accurately perceiving the physical environment. Meanwhile, transfer learning is also widely used in semantic segmentation since it requires dense pixel-wise manual labeling which is too expensive to obtain a large amount of training data for supervision. Thus it is inherent to apply transfer learning in semantic segmentation to ease the demand of labeled data. \par

 To our knowledge, almost all of the transferability metrics are targeting to the image classification problem, and there is no specific work to thoroughly investigate the transferability  estimation of semantic segmentation task. Although the semantic segmentation task can be  considered as a classification problem per pixel, it is difficult to directly apply existing transferability metrics due to the high dimensional output of the segmentation model. Therefore, in this paper, we propose to extend existing transferability metric OTCE (Optimal Transport based Conditional Entropy) score~\cite{OTCE} to be capable of estimating the transferability of semantic segmentation model. Generally, OTCE score builds the correspondences (joint probability distribution) between source and target data via solving an Optimal  Transport (OT)~\cite{kantorovich1942translocation} problem, and then compute the Conditional Entropy (CE) between source and target label as the \textit{task difference} to describe transferability. In practice, this framework can efficiently work under the sample size of $10^4$ which satisfies the data scale of most image classification tasks. However, the entity of segmentation task is \textit{pixel} instead of \textit{image}, so that the sample size of a semantic segmentation dataset will dramatically increase to $10^7$, which is unsolvable under limited computation resources, e.g, personal computer. Therefore, we propose to sample a set of pixels from the source and target datasets respectively for computing the OTCE score, and take the expectation over K repetitions as the final transferability score. \par
 
 We conduct extensive experiments under the \textit{intra-dataset} and \textit{inter-dataset} transfer settings for evaluation. For intra-dataset transfer setting, we successively take each city from the Cityscapes~\cite{cityscapes} dataset as the target task, and others are considered as source tasks. For inter-dataset transfer setting, we transfer six source models with different architectures pretrained on BDD100K~\cite{bdd100k} and GTA5~\cite{gta5} datasets respectively to Cityscapes. Results clearly demonstrate that the OTCE score is highly correlated with the real transfer performance, indicating that OTCE score can also be applied for the transferability estimation of semantic segmentation tasks. 

\section{Method}
In this section, we first define the transferability problem of semantic segmentation task. And then we will introduce the computation process of OTCE score applying in semantic segmentation. 

\subsection{Transferability Problem}
Formally, suppose we have a source dataset $D_s = \{(x^i_s, y^i_s)\}_{i=1}^m \sim P_s(x,y)$ and a target dataset $D_t = \{(x^i_t, y^i_t)\}_{i=1}^n \sim P_t(x,y)$, where $x^i_s, x^i_t \in \mathcal{X}$ and $ y^i_s \in \mathcal{Y}_s, y^i_t \in \mathcal{Y}_t$. Here $x$ and $y$ represent the input image and the label mask of size $H\times W$ respectively. Meanwhile, $P(x_s) \neq P(x_t)$ and $\mathcal{Y}_s \neq \mathcal{Y}_t$ indicate different domains and tasks respectively. In addition, we have a source model $\theta_s: \mathcal{X}\rightarrow \mathcal{P}(\mathcal{Y}_s)$ trained on the source dataset $D_s$, where $\mathcal{P}(\mathcal{Y}_s)$ is the space of all probability distributions over $\mathcal{Y}_s$. \par

For neural network based transfer learning, we usually adopt the \textit{finetune} method to transfer the knowledge learned from the source task to the target task. Specifically, we initialize the target model $\theta_t: \mathcal{X}\rightarrow \mathcal{P}(\mathcal{Y}_t)$ from the source weights, and then continue the training process on the target training data. Note that when $|\mathcal{Y}_s| \neq |\mathcal{Y}_t|$, we need to modify the final layer of the target model to ensure the mapping $\mathcal{X}\rightarrow \mathcal{P}(\mathcal{Y}_t)$. Then the empirical transferability can be defined as:
\begin{definition}
	The empirical transferability from source task $S$ to target task $T$ is measured by the expected log-likelihood of the retrained $\theta_t$ on the testing set of target task:
	\begin{equation}
	\mathrm{Trf}(S \rightarrow T) = \mathbb{E} \left[ \mathrm{log}\ P(y_t|x_t; \theta_t) \right] 
	\label{eq:true-transferability}
	\end{equation}
	which indicates how good the transfer performance is on target task $T$. \cite{NCE, OTCE} 
\end{definition}\par

Although the empirical transferability accurately reveals the \textit{ground-truth} of the transfer performance, it is computationally expensive to obtain. Consequently, we hope to use the analytical transferability metric, e.g. OTCE~\cite{OTCE}, to efficiently approximate the empirical transferability. 

\subsection{OTCE for Semantic Segmentation Task}

\begin{figure}[h]
	\centering
	\def\svgwidth{0.8\linewidth}
	\executeiffilenewer{images/OTCE.svg}{images/OTCE.pdf}%
	{inkscape -z -D --file=images/OTCE.svg %
		--export-pdf=images/OTCE.pdf --export-latex}%
	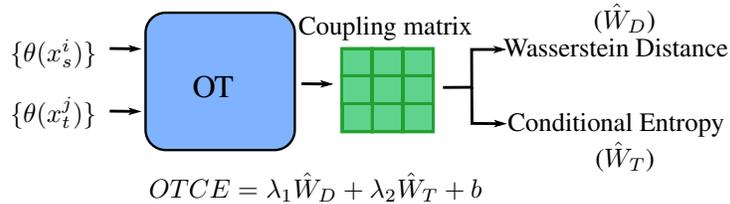%

	\caption{Illustration of OTCE score~\cite{OTCE}.}
	\label{fig: illustration of OTCE}
\end{figure}

OTCE score is a unified framework (shown in Figure \ref{fig: illustration of OTCE}) which characterizes the \textit{domain difference} and the \textit{task difference} between source and target tasks, and uses the linear combination of domain difference and task difference to describe transferability. Specifically, the OTCE score first estimates the joint probability distribution $\hat{P}(X_s, X_t)$ of source and target input instances via solving an Optimal Transport (OT) problem~\cite{kantorovich1942translocation}, which also produces the Wasserstein distance (domain difference). Then based on $\hat{P}(X_s, X_t)$, we can obtain $\hat{P}(Y_s, Y_t)$ and $\hat{P}(Y_s)$ for calculating the Conditional Entropy $H(Y_t|Y_s)$ (task difference).\par 

As the original OTCE requires several auxiliary tasks with known empirical transferability for learning the linear combination of domain difference and task difference, we trade-off accuracy for a more practical implementation, i.e., only take the task difference for describing transferability.\par

Formally, we first use the source model $\theta_s$ to extract feature maps $\{\theta_s(x^i_s)\}^m_{i=1}$ from the source dataset $D_s$ and $\{\theta_s(x^i_t)\}^n_{i=1}$ from the target dataset $D_t$. Then we take the feature vector $f$ and the label $y$ of each pixel to construct pixel-wise datasets, denoted as $D^{pix}_s=\{(f^i_s,y^i_s)\}^{N_s}_{i=1}$ for the source task and $D^{pix}_t=\{(f^i_t,y^i_t)\}^{N_t}_{i=1}$ for the target task, where $N_s=m \times H \times W$ and $N_t=n \times H \times W$ represent the number of pixels in the source and target datasets respectively. Then we define the OT problem with an entropic regularizer~\cite{cuturi2013sinkhorn} in our setting to faciliate the computation:   

\begin{equation}
\begin{aligned}
OT(D^{pix}_s, D^{pix}_t)  \triangleq  \mathop{\min}\limits_{\pi \in \Pi(D^{pix}_s, D^{pix}_t)}  \sum_{i,j=1}^{N_s,N_t}  c(f^i_s,f^j_t)\pi_{ij} + \epsilon H(\pi), 
\end{aligned}
\label{eq: OT definition with entropy}
\end{equation}
where $c(\cdot, \cdot) = \| \cdot - \cdot \| ^2_2$ is the cost metric, and $\pi$ is the coupling matrix of size $N_s\times N_t$, and $H(\pi)=-\sum_{i=1}^{N_s} \sum_{j=1}^{N_t} \pi_{ij}\log\pi_{ij}$ is the entropic regularizer with $\epsilon=0.1$. The OT problem above can be solved efficiently by the Sinkhorn algorithm~\cite{cuturi2013sinkhorn} to produce an optimal coupling matrix $\pi^*$. Then we can compute the empirical joint probability distribution of source and target label sets, and the marginal probability distribution of source label set, denoted as: 
\begin{equation}
\hat{P}(y_s,y_t) = \sum_{i,j: y^i_s=y_s, y^j_t=y_t} \pi_{ij}^*,
\label{eq: joint distribution}
\end{equation}

\begin{equation}
\hat{P}(y_s) = \sum_{y_t \in \mathcal{Y}_t} \hat{P}(y_s,y_t).
\label{eq: marginal distribution}
\end{equation}
Thus the \textit{task difference} $W_T$ can be computed as the Conditional Entropy (CE) between the source and target label.
\begin{equation}
\begin{aligned}
W_T & = H(Y_t|Y_s) = H(Y_s,Y_t) - H(Y_s)\\
& = - \sum_{y_t \in \mathcal{Y}_t} \sum_{y_s \in \mathcal{Y}_s} \hat{P}(y_s,y_t)\log \frac{\hat{P}(y_s,y_t)}{\hat{P}(y_s)}.
\end{aligned}
\label{eq: task difference}
\end{equation}
Finally, we take the negative task difference as the OTCE score for describing transferability.
\begin{equation}
OTCE = - W_T.
\label{eq: OTCE score}
\end{equation}

\subsection{Implementation}

Recall that the semantic segmentation task can be regarded as the classification problem per pixel and a dataset may contain more than $10^7$ pixels (100 images of size $1024 \times 512$) making it difficult to compute the OTCE score under limited computation resources, e.g., personal computer. Therefore, we propose to randomly sample N pixels from the source and target pixel sets respectively for computing OTCE score, and then take the expectation over K repetitions as the final transferability score. Specifically, the computation process is decribed in Algorithm \ref{alg:OTCE}. In our experiments, we let $N=10,000$ and $K=10$, and they can be adjusted according to different settings. 

\begin{algorithm}[ht]
	\caption{OTCE for Semantic Segmentation Task}
	\label{alg:OTCE}
	\begin{algorithmic}[1]
		{\footnotesize
			\REQUIRE $D^{pix}_s = \{(f_s^i, y_s^i)\}_{i=1}^{N_s}$: The source pixel set.
			\REQUIRE $D^{pix}_t = \{(f_t^i, y_t^i)\}_{i=1}^{N_t}$: The target pixel set.
			
			\STATE Let $k = 0$;
			\STATE Let $score = 0$;
			
			\FOR{$ k < K$}
			\STATE $\hat{D}^{pix}_s = RandomlySample(D^{pix}_s, N)$; (Randomly sample N pixels.)
			\STATE $\hat{D}^{pix}_t = RandomlySample(D^{pix}_t, N)$; (Randomly sample N pixels.)
			\STATE $score += OTCE(\hat{D}^{pix}_s, \hat{D}^{pix}_t)$;
			\ENDFOR
			\STATE \textbf{Return}  $score / K$
			
		}
	\end{algorithmic}
\end{algorithm}

\begin{figure}[h]
	\centering
	\includegraphics[width=1.0\linewidth]{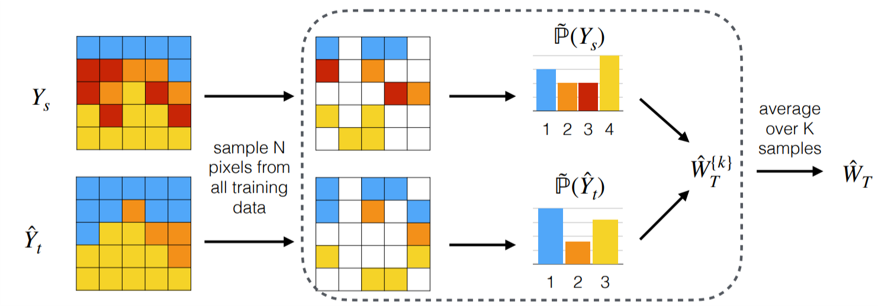}
	\caption{A toy illustration of the OTCE for semantic segmentation task.}
	\label{fig:toy example}
\end{figure}
\section{Experiment}
We evaluate the effectiveness of OTCE score applied in semantic segmentation task under various transfer settings. First, we investigate the performance under the \textit{intra-dataset} transfer setting, i.e., source task and target task come from the same dataset but different sub-domains. We successively take the city from the Cityscapes~\cite{cityscapes} as the target task, and transfer the source models pretrained on other cities to it. For the \textit{inter-dataset} transfe setting, we also take the city from the Cityscapes as the target task, but transfer from the source models trained on the BDD100K~\cite{bdd100k} and GTA5~\cite{gta5} dataset, where source models have different architectures.

\subsection{Evaluation on Intra-dataset Transfer Setting}

Cityscapes is a large-scale semantic understanding dataset captured in street scenes from different cities with diverse season, time, background and weather. Consequently, we consider each city to be a sub-domain. They annotate 21 cities with dense semantic label mask of 34 classes. We successively take one city as the target task, and remains are source tasks. All source models are defined in a UNet~\cite{unet} architecuture, and trained on the full source training data. As the transfer learning usually adopted in the few-shot case, we only randomly sample 20 target training instances for finetuning the source model. We also use Pearson correlation coefficient like \cite{NCE, LEEP, OTCE} to quantitatively evaluate the correlation between the transfer accuracy and the transferability score.\par

\begin{figure}[h]
	\centering
	\includegraphics[width=0.8\linewidth]{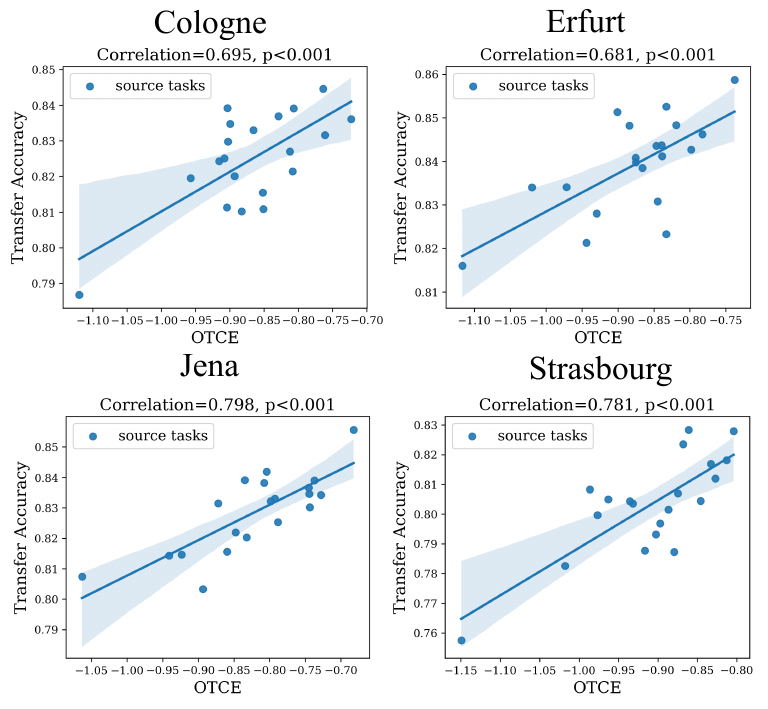}
	\caption{Correlation between the transfer accuracy and the OTCE score under the intra-dataset transfer setting, where target tasks are four sub-domains (cities) including \textit{Cologne, Erfurt, Jena} and \textit{Strasbourg} selected from the Cityscapes. Each target task has 20 source models trained on other sub-domains of Cityscapes.  }
	\label{fig:intra-dataset}
\end{figure}

Results shown in Figure \ref{fig:intra-dataset}  demonstrate that our predicted OTCE score highly correlates with the transfer performance achieving up to 0.768 correlation score, which suggests that it is reliable to select the highly transferable source task according to the OTCE score. 

\subsection{Evaluation on Inter-dataset Transfer Setting}

\begin{figure}[h]
	\centering
	\includegraphics[width=0.8\linewidth]{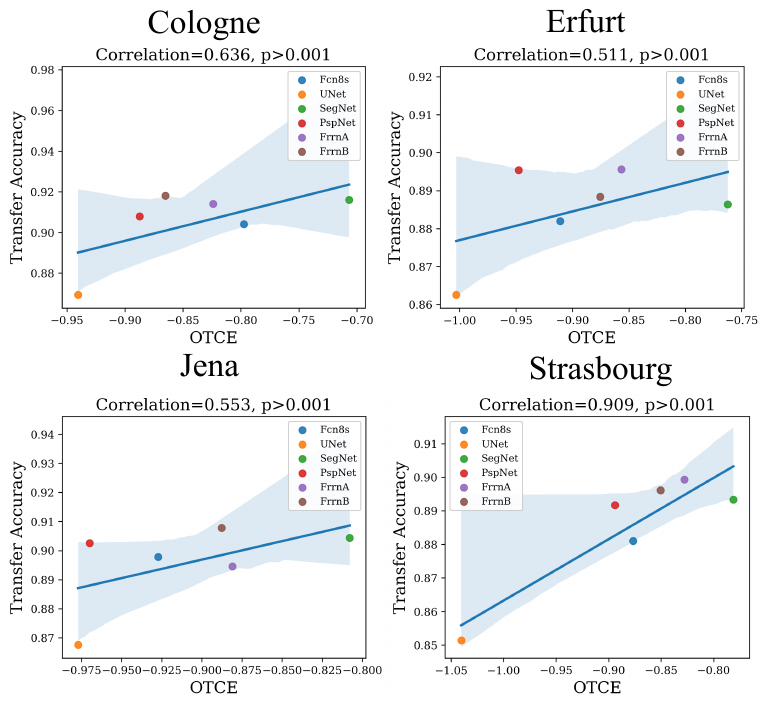}
	\caption{Correlation between the transfer accuracy and the OTCE score under the inter-dataset transfer setting, where six source models are trained on \textbf{BDD100K}. Target tasks are four sub-domains (cities) including \textit{Cologne, Erfurt, Jena} and \textit{Strasbourg} selected from the Cityscapes. }
	\label{fig:inter-dataset-bdd}
\end{figure}

\begin{figure}[h]
	\centering
	\includegraphics[width=0.8\linewidth]{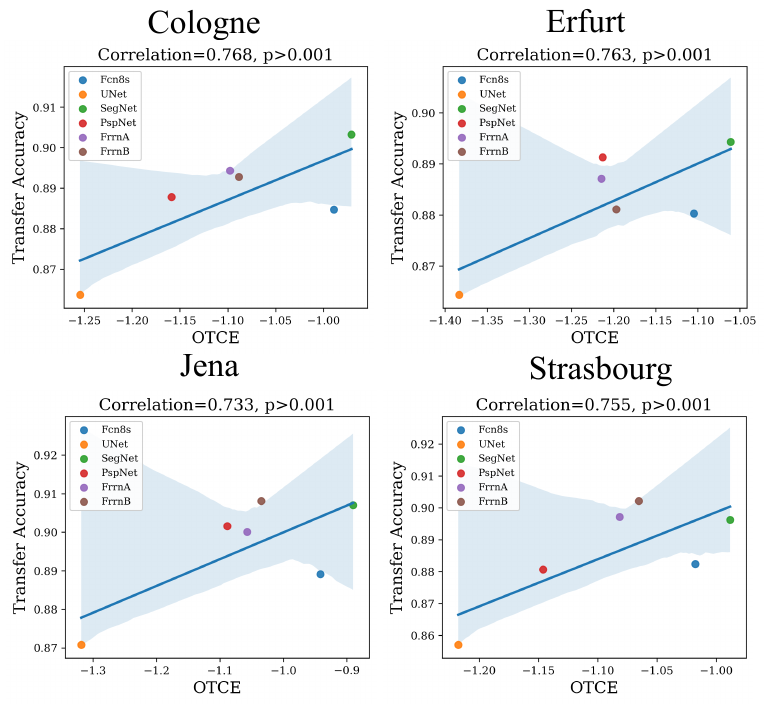}
	\caption{Correlation between the transfer accuracy and the OTCE score under the inter-dataset transfer setting, where six source models are trained on \textbf{GTA5}. Target tasks are four sub-domains (cities) including \textit{Cologne, Erfurt, Jena} and \textit{Strasbourg} selected from the Cityscapes. }
	\label{fig:inter-dataset-gta}
\end{figure}

We introduce another two datasets BDD100K and GTA5 for further investigating the performance under the more challenging inter-dataset transfer setting. Specifically, we train six models including Fcn8s~\cite{fcn8s}, UNet~\cite{unet}, SegNet~\cite{segnet}, PspNet~\cite{pspnet}, FrrnA~\cite{frrn} and FrrnB~\cite{frrn} on BDD100K and GTA5 respectively, and then we transfer those source models to the sub-domains of Cityscapes. Note that BDD100K is the real captured data, but GTA5 is generated from the computer game. Thus the domain gap between Cityscapes and GTA5 is larger than that between Cityscapes and BDD100K. The transferability estimation performance is also evaluated by Pearson correlation coefficient. Figure \ref{fig:inter-dataset-bdd}, \ref{fig:inter-dataset-gta} visualize the correlation between the transfer accuracy and the OTCE score under the configuration of transferring from BDD100K and GTA5 respectively.  We can see that the OTCE score can be a good indicator for selecting highly transferable source model under the inter-dataset transfer setting. And transferring from BDD100K achieves higher transfer accuracy than that from the GTA5 dataset, which suggests that larger domain gap may worsen the transfer performance. 

\section{Conclusion}

We thoroughly investigate the transferability estimation problem of semantic segmentation task using the OTCE score under various transfer settings, including the intra-dataset transfer and the inter-dataset transfer. As the original OTCE score is unable to deal with large-scale data making it difficult to apply in pixel-wise classification (semantic segmentation) problem, we propose a random sampling strategy, i.e., sample N pixels for computing the OTCE score and take the expected OTCE score over K repetitions. Experiments show that the OTCE score can also be a good indicator for selecting the highly transferable source task or source model architecture on semantic segmentation task.

\bibliographystyle{ieee_fullname}
\bibliography{egbib}

\end{document}